\documentclass[sigconf,screen,nonacm]{acmart}

\usepackage[noend]{algorithmic}
\usepackage{algorithm}
\usepackage{booktabs} 

\newcommand{\algrule}[1][.2pt]{\par\vskip.2\baselineskip\hrule height #1\par\vskip.2\baselineskip}

\AtBeginDocument{%
  \providecommand\BibTeX{{%
    \normalfont B\kern-0.5em{\scshape i\kern-0.25em b}\kern-0.8em\TeX}}}

\setcopyright{acmcopyright}
\copyrightyear{2018}
\acmYear{2018}
\acmDOI{10.1145/1122445.1122456}

\acmConference[Woodstock '18]{Woodstock '18: ACM Symposium on Neural
  Gaze Detection}{June 03--05, 2018}{Woodstock, NY}
\acmBooktitle{Woodstock '18: ACM Symposium on Neural Gaze Detection,
  June 03--05, 2018, Woodstock, NY}
\acmPrice{15.00}
\acmISBN{978-1-4503-XXXX-X/18/06}

\begin{document}

\title[Ensemble Learning based on Classifier Prediction Confidence and CLPSO for polyp localisation]{Ensemble Learning based on Classifier Prediction Confidence and Comprehensive Learning Particle Swarm Optimisation for polyp localisation}





\author{Truong Dang}
\affiliation{%
  \institution{Robert Gordon University, School of Computing}
  \city{Aberdeen}
  \country{UK}}
\email{dangmanhtruong@gmail.com}

\author{Thanh Nguyen}
\affiliation{%
  \institution{Robert Gordon University, School of Computing}
  \city{Aberdeen}
  \country{UK}}
\email{t.nguyen11@rgu.ac.uk}
\author{John McCall}                                
\affiliation{%
  \institution{Robert Gordon University, School of Computing}
  \city{Aberdeen}
  \country{UK}}
\email{j.mccall@rgu.ac.uk}
\author{Alan Wee-Chung Liew}                                
\affiliation{%
  \institution{Griffith University, School of Information and Communication Technology}
  \city{Queensland}
  \country{Australia}}
\email{a.liew@griffith.edu.au}
\renewcommand{\shortauthors}{Author et al.}

\begin{abstract}

Colorectal cancer (CRC) is the first cause of death in many countries. CRC originates from a small clump of cells on the lining of the colon called polyps, which over time might grow and become malignant. Early detection and removal of polyps are therefore necessary for the prevention of colon cancer. In this paper, we introduce an ensemble of medical polyp segmentation algorithms. Based on an observation that different segmentation algorithms will perform well on different subsets of examples because of the nature and size of training sets they have been exposed to and because of method-intrinsic factors, we propose to measure the confidence in the prediction of each algorithm and then use an associate threshold to determine whether the confidence is acceptable or not. An algorithm is selected for the ensemble if the confidence is below its associate threshold. The optimal threshold for each segmentation algorithm is found by using Comprehensive Learning Particle Swarm Optimization (CLPSO), a swarm intelligence algorithm. The Dice coefficient, a popular performance metric for image segmentation, is used as the fitness criteria. Experimental results on two polyp segmentation datasets MICCAI2015 and Kvasir-SEG confirm that our ensemble achieves better results compared to some well-known segmentation algorithms.

\end{abstract}




\settopmatter{printfolios=true}


\maketitle


\section{Introduction}

Colon cancer or colorectal cancer (CRC) is  one of the most common causes of death worldwide, with around 1,360,000 newly diagnosed cases and 694,000 mortality cases each year \cite{ferlay_cancer_2015}. CRC arises from adenomatous polyps (or adenomas), which are growths of glandular tissue originating from the colonic mucosa. These polyps are initially benign, but over time they might become malignant and spread to other organs such as the liver and lung, eventually resulting in death \cite{miccai_2015_paper}. A crucial step in CRC prevention is the detection of polyps before they turn malignant or are at the early stage of cancer. The procedure for doing this is called colonoscopy. In this stage only the most superficial colon layers are involved without any deep invasion. Once the polyps are identified, the clinicians can then perform surgical removal. Even though colonoscopy is considered the gold-standard for colon screening, other alternatives such as CT colonography or wireless capsule endoscopy (WCE) are also used. These methods are highly dependant on the clinician's skills, and if misinterpretation of data is taken into consideration, the accuracy rate decreases sharply  and  the  duration of early  detection is prolonged \cite{waite_systemic_2017}. Therefore it is crucial to automate the process of early polyp detection and localisation.

The polyp can be detected and localised automatically from images based on segmentation algorithms. Segmentation refers to the process in which an image is partitioned into  a number of segments which delineate different kind of objects. Before the rise of deep learning in 2012 \cite{krizhevsky_imagenet_2012}, most successful segmentation techniques extracted hand-crafted features which are then used as input to a machine learning method. Generally, the performance of these systems is limited because the hand-crafted features were not representative enough for real-world situations. Since 2012, there have been many applications of deep learning to segmentation. A notable example is Fully Connected Network (FCN) \cite{fcn_paper_2015} which is created by using a pretrained deep network for image classification as the backbone and then the final Fully Connected (FC) layer is converted into upsampling layers to produce dense pixel-level output for segmentation. 

Deep learning can potentially be applied to polyp segmentation for early colorectal cancer diagnosis. However, compared with other problems like image classification in which there are many datasets having millions of examples, such as ImageNet \cite{imagenet_cvpr_2009}, the amount of publicly available medical images is still limited. Considering that the breakthrough of deep learning was achieved by training on ImageNet, large visual database with more than 14 million images \cite{krizhevsky_imagenet_2012}, this means that deep learning models for medical images are still not exploited to their full potential. Another problem is that deep learning models generally require  careful parameter tuning to achieve good results. These shortcomings create challenges in choosing a suitable and robust deep learning model for clinical applications. One solution for these challenges is to exploit the strength of multiple segmentation algorithms to provide an improved result.

Ensemble learning is a popular technique in which a number of machine learning methods are combined to create a collaborated decision. However, it is observed that not any combination gives the desired results. The presence of some methods may downgrade the ensemble performance and they should be removed from the ensemble. The idea of this paper is based on the real-life observation that when a committee of experts consults on a problem, each of them usually has different background and level of expertise. If an expert is known to be very knowledgeable in a field, his/her recommendation would be trusted even though he/she might not be sure about the current recommendation. In contrast, if an expert is not knowledgeable about the issue being discussed then we would not trust his/her recommendation even if he is very sure of it. We apply this idea to select the optimal subset of deep segmentation algorithms for polyp segmentation. The expertise level of each algorithm is encoded by using a threshold. The confidence of the prediction of each algorithm is measured and then compared with the corresponding threshold to determine whether this algorithm should be included in the ensemble. We propose using Shannon entropy to measure confidence in the prediction. The optimal threshold for each segmentation algorithm is found by maximizing the Dice coefficient, a popular performance metric for image segmentation, using Comprehensive Learning Particle Swarm Optimisation (CLPSO), a swarm intelligence algorithm. 

The paper is organised as follows. In section 2, we provide a brief review of the existing approaches relating to polyp  segmentation, ensemble learning, PSO, and Comprehensive Learning. Our proposed ensemble is introduced in section 3. The details of experimental studies on two polyp segmentation datasets are described in section 4. Finally, the conclusion is given in section 5. 

\section{Background and Related Work}
\subsection{Polyp segmentation}
\begin{figure}
    \begin{center}
        \includegraphics[width=0.3\textwidth, height=0.3\textwidth]{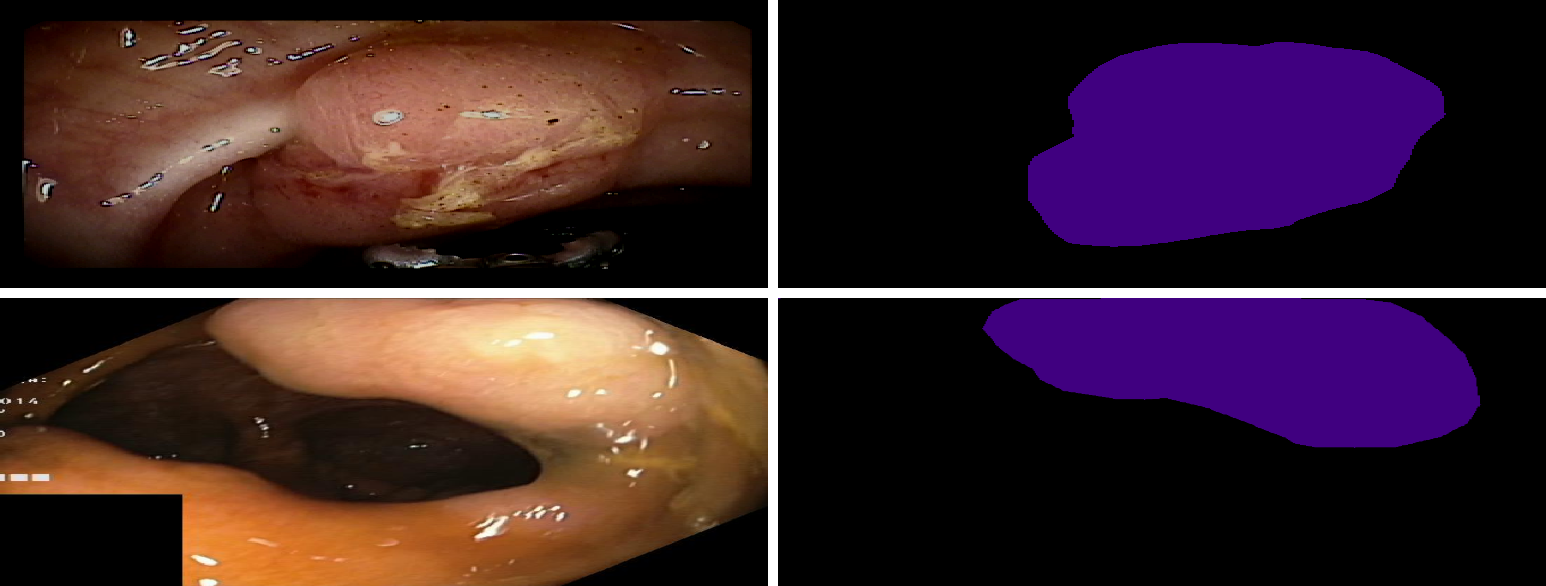}
    \end{center}
    \caption{Example polyp images from the MICCAI2015 and Kvasir-SEG dataset (left) and their corresponding ground truths (right). The first row is from the MICCAI2015 dataset and the second row is from the Kvasir-SEG dataset. The purple areas indicate the polyps, while the black area indicates the background.}
    \label{fig:polyp_example_image_and_mask}
\end{figure}
Before the rise of deep learning, the majority of works on polyp segmentation relied on hand-crafting low-level image processing methods to obtain candidate polyp boundaries. For example, \cite{zhu_improved_2011}, used the Knutsson mapping method to provide curvature estimations of the polyps boundaries compared to previous methods, while \cite{hwang_polyp_2007} combined shape fitting with curvature analysis in the segmentation of polyps. Since its success in image classification in 2012, deep learning has been widely applied to segmentation. One of the first successful architectures was Fully Convolutional Network (FCN) \cite{fcn_paper_2015}. This architecture uses an existing classification network, such as VGG16 \cite{vgg16_paper_2015}, as the backbone and replaces the fully connected layers with upsampling layers to produce pixel-level segmentation result. There have also been deep networks specifically designed for the segmentation of medical images. A notable example is UNet \cite{unet_olaf_2015}, a deep segmentation network designed for the problem of segmentation of neuronal structures in electron microscopic stacks. Building upon FCN, the authors combined high resolution features from the convolutional layers with the upsampled output, which facilitates more precise segmentation based on this information. An important contribution of this method is that in the upsampling part there is also a large number of feature channels which allow the network to propagate context information to successive layers. The network is therefore largely symmetric. Other notable examples are LinkNet \cite{linknet_paper_2017} which takes the sum of the upsampled output and the corresponding features in the convolutional path, and Feature Pyramid Network (FPN) \cite{fpn_paper_2017} which uses the concatenation of features of all levels in the upsampling part to help with the final prediction.

Recently, \cite{jia_polyp_2020} introduced a deep two-stage architecture that includes an advanced method that consisted of residual learning and feature pyramids. In addition, the architecture uses a feature-sharing strategy for transferring semantic information during training. Another approach is \cite{feng_polyp_2020} which proposed Stair-Shape Network (SSN) for real-time polyp segmentation. The architecture utilises four blocks for feature extraction at the encoder stage. In each block, there is a Dual Attention Module and a final Multi-scale Fusion Module is used to fuse the features at each scale. Strong data augmentation and auxiliary losses are used to improve segmentation results. \cite{sun_polyp_2019} introduced a novel deep learning framework based on UNet \cite{unet_olaf_2015} for the colorectal polyp segmentation. The authors improved on the design of UNet by introducing dilated convolution to learn high-level semantic features without a reduction in resolution. The decoder stage is simplified by combining multi-scale semantic features. The authors also developed post processing techniques to improve colorectal polyp detection performance. There are few medical datasets for polyp detection and localization which includes ground truth that meets medical standard. For example, CVC-ColonDB \cite{tajbakhsh_polyp_2016} consists of 15 short colonoscopy video sequences, containing a total of 1200 frames. However, only 300 frames are annotated. In 2015, the Automatic Polyp Detection sub-challenge, conducted as part of the Endoscopic Vision Challenge (\url{http://endovis.grand-challenge.org}) at the Medical Image Computing and Computer Assisted Intervention (MICCAI) was organised \cite{miccai_2015_paper}. The purpose of this competition was to assess the clinical applicability of segmentation methods when faced with technical and clinical challenges reported in the literature. Several datasets were used for polyp localization in images and videos. The authors reported the best results by competitors, consisting of one hand-crafted method, three deep learning methods, and three hybrid methods. Figure \ref{fig:polyp_example_image_and_mask} shows an example image and its corresponding ground truth.

\subsection{Ensemble learning and Ensemble selection}
Ensemble learning is a popular machine learning technique in which multiple learners i.e. classifiers are combined to improve the overall performance. Typically, ensemble systems are built by either training a learning algorithm on multiple training sets generated from the original training data or training different learning algorithms on the original training data to generate the ensemble \cite{nguyen_projection_2019, nguyen_variational_2016}. Afterwards, a combining method is then applied to the predictions of the generated classifiers for the final decision. There are some techniques concerning the combining methods. Nguyen et al. \cite{nguyen_projection_2019} searched for the weights of classifiers in the combining by minimizing the distance between these combinations computed on the training data and the class label of training observations given in the binary form. Sen et al. \cite{sen_regularization_2013} searched for the combining weights to minimise the hinge loss function of the combination and the training labels. The approach of Zhang and Zhou \cite{zhang_semideﬁnite_2006} used linear programming to search for the combining weights. Pacheco et al. \cite{pacheco_dynamic_2020} modelled the output probabilities as a Dirichlet distribution and optimised the weights of classifiers using a loss function based on Mahalanobis distance.

Meanwhile, based on the observation that the presence of some classifiers might lower the performance of the ensemble, there have been many research efforts into \textit{Ensemble Selection} (ES) (also known as ensemble pruning) which aims to select a subset of classifiers which is competitive to the whole ensemble. There are two approaches to ensemble selection: static or dynamic approach. The static approach selects a subset of classifiers during the training phase and uses it for the testing phase. This approach limits the flexibility of the selection procedure \cite{nguyen_confidence_2020}. In contrast, the dynamic approach selects a different subset of classifiers for each test instance. The static approach can be further divided into ordering-based methods and optimisation-based methods. The ordering-based methods try to order the classifiers according to ranking criteria e.g. validation error \cite{Margineantu_pruning_1997} or margin \cite{martinez_aggregation_2004}, among which only the top classifiers are selected. Optimisation-based methods formulate ensemble selection as an optimisation problem 
which can be solved by heuristic optimisation or mathematical programming \cite{nguyen_confidence_2020}. For example, Ant Colony Optimisation (ACO) was used in \cite{chen_applying_2014} to find the optimal set of classifiers and combining method in the ensemble systems. In \cite{nguyen_genetic_2014}, the authors introduced an encoding for both the classifiers and the features in a single chromosome and used a Genetic Algorithm (GA) to simultaneously search for the optimal set of classifiers and the associated features. In contrast, in the dynamic approach, a classifier is selected based on its performance in a local region of the feature space called Region of Competence (RoC) \cite{nguyen_confidence_2020}. A comparative review of dynamic methods can be found in \cite{britto_dynamic_2014}. 

Evolutionary Computation (EC) refers to an area of computational intelligence which uses ideas from biological evolution as inspiration to solve computational problems \cite{mitchell_evolutionary_1999}. There has been many works on applying EC to ensemble learning. The main rationale for this approach is that  they provide collections of hypotheses, extracted using many runs or within a single run using diversity enforcing heuristics \cite{jong_ensemble_2004}. \cite{alberto_genetic_2016} introduced a multi-level ensemble of Least Square Support Vector Machine \cite{burges_tutorial_1998} having three levels: input space, the base components and the combining block of the components responses. Genetic algorithm (GA) was used to optimize the ensemble. \cite{jeongju_fault_2019} used GA to find the optimal ensemble of fault localisation ranking models and the results on 389 real-world faults in a popular benchmark indicated the effectiveness of the proposed method. An extension of Genetic Programming (GP) was used in \cite{folino_ensemble_2006} to optimize an ensemble of predictors using voting classification schemes based on bagging and boosting techniques. The experiments showed that the tree size is reduced while accuracy and execution time are improved. To reduce computational complexity of running GP training methods to obtain the models, \cite{dick_evolving_2018} introduced a method based on spatial structure with bootstrap elitism. \cite{bhowan_evolving_2011} used GP to build a classifier ensemble with unbalanced data by optimizing a multi-objective problem, with the two objectives being the minority and majority class accuracy. There are also other methods based on other EC methods such as Particle Swarm Optimisation (PSO) \cite{ripon_efficient_2020}.  

It is widely recognized that diversity is an important factor in the design of ensemble methods \cite{kuncheva_measures_2003}. Although there is no precise definition of ensemble diversity, there has been many proposed measures to quantify the diversity of an ensemble. Diversity measures can be divided into pairwise diversity and non-pairwise diversity \cite{gomes_survey_2017}. Pairwise diversity is based on the difference of predictions between a pair of classifiers, and the average of pairwise differences is used as the overall diversity. An example of pairwise diversity measure is Q-statistic which is derived as the equivalent of the correlation coefficient for binary valued measurements \cite{kuncheva_limits_2003}. Other notable pairwise-based measures include K-statistic \cite{dietterich_experimental_2000} and pairwise failure crediting \cite{chandra_ensemble_2006}. In contrast, nonpairwise diversity directly measures a set of classifiers based on variance, entropy or other global metrics \cite{shenkai_ensemble_2015}. A notable example is  Kohavi-Wolpert variance which measures the variability of predicted class labels by each classifier \cite{kuncheva_measures_2003}. Another approach is correlation penalty function in which  the diversity of each ensemble member is measured against the entire ensemble \cite{liu_ensemble_1999}. 

\subsection{Particle Swarm Optimisation and Comprehensive Learning}

Particle Swarm Optimisation (PSO) is a swarm-based algorithm inspired by the emergent motion of a flock of birds searching for food \cite{kennedy_pso_1995}. Each particle performs local exploitation simultaneously with the global exploration by the whole swarm. In PSO, each swarm member, called a particle, represents a candidate solution
 in the search space. The global optimum is regarded as the location of food. Each particle is associated with a fitness value and a velocity to adjust its flying direction according to the best experiences of the swarm to search for the global optimum in the search space. Since its introduction, PSO has attracted a high level of interest \cite{pso_application_paper_2004} and therefore has seen many research efforts into improving its performance. For example \cite{shi_modified_1998} introduced an inertia weight term to balance the global and local search abilities. \cite{kennedy_particle_2002} analysed the convergence properties of PSO and designed a variant with constriction factor which guarantees the convergence and improves the convergence velocity. Another direction is to design different types of PSO topologies. \cite{hu_optimization_2002} used a dynamic neighborhood where $m$ closest particles are chosen to be the new neighborhood in each generation. \cite{parsopoulos_upso_2019} created a combination of the global version and local version called unified particle swarm optimiser (UPSO). Some researchers also investigated the hybridisation of PSO with other search techniques, such as evolutionary operators like crossover or mutation \cite{angeline_selection_1998}. Although many variants of PSO have been designed, the main deficiency of PSO is still premature convergence \cite{liang_comprehensive_2006}. In the original PSO, each particle only learns from its best position so far ($\textit{pbest}$) and global best position ($\textit{gbest}$) which makes it converge quickly. However, if the $\textit{gbest}$ gets trapped in a local optimum then other particles might be attracted to it, leading to premature convergence. \cite{liang_comprehensive_2006} introduced Comprehensive Learning PSO (CLPSO) to mitigate this problem by having each particle learn from all particles' local best position. In this method, each particle learns from exemplars which are chosen from the previous best positions of all other particles and each dimension of a particle can potentially learn from a different exemplar. The authors compared CLPSO with eight PSO variants on 16 benchmark problems and found that the new strategy makes use of the information in swarm more effectively to generate better quality solutions.

\section{Proposed method}

Let $\textbf{D}$ be the training set of $N$ observations $\{(\textbf{I}_n,\textbf{Y}_n )\}_{n=1}^N$ where $\textbf{I}_n$ is the $n^{th}$ training image, and $\textbf{Y}_n$ is the corresponding ground truth. The ground truth $\textbf{Y}_n$ has the same size as $\textbf{I}_n$ in which each position denotes the class label of the corresponding image pixel. Each class label belongs to a set of labels $\mathcal{Y}=\{y_m\}_{m=1}^M$ i.e.  $\textbf{Y}_n (i,j) \in \mathcal{Y}(1 \leq i \leq W, 1 \leq j \leq H)$. Let $\mathcal{\textbf{K}}=\{\mathcal{K}_k\}_{k=1}^K$  be the set of $K$ segmentation algorithms and each learning algorithm $\mathcal{K}_k$ trains the segmentation model $\mathcal{C}_k$ on the training data $\textbf{D}$. For an image $\textbf{I}$, let $P_{k,m}(\textbf{I}(i,j))$ denote the prediction probability by the model associated with $\mathcal{K}_k$ that the pixel $\textbf{I}(i,j)(1 \leq i \leq W, 1 \leq j \leq H)$ belongs to class $y_m$. There are several constraints on $\{P_{k,m}(\textbf{I}(i,j))\}$ as $0 \leq P_{k,m}(\textbf{I}(i,j)) \leq 1$ and  $\sum_{k=1}^{K}P_{k,m}(\textbf{I}(i,j))=1$ for each $m$. In ensemble learning, the prediction probabilities $\{P_{k,m}(\textbf{I}(i,j))\}$ of the $K$ models are combined to obtain the final prediction. 

In ensemble learning, usually the predictions from all methods are used for combination to create the final prediction.  However, it is possible that the presence of some methods degrades the ensemble performance. There have been many research efforts into Ensemble Selection (ES) to select a subset of methods which performs competitively to or even better than the whole ensemble. Our idea is based on the observation in real-life when consultation from an expert committee is required. An expert which is experienced in a particular field should be trusted when working on this field even though he/she is not entirely sure about his/her recommendation. In contrast, when an expert is not knowledgeable about the current problem his/her opinion should only be regarded even though he/she is completely sure. Applying this idea to our problem, it can be seen that for optimal selection of deep segmentation algorithms, each algorithm should have a particular evaluation criteria for selection into the ensemble. In this study, we introduce a novel ensemble selection method in order to increase ensemble performance. We compute the $\textit{Shannon entropy}$ of the prediction by $k^{th}$ algorithm on pixel $\textbf{I}(i,j)$  as follows:
\begin{equation}
    \label{eq:entropy_definition}
    \textbf{E}_k(\textbf{I}(i,j))=- \sum_{m=1}^{M}P_{k,m}(\textbf{I}(i,j))*log(P_{k,m}(\textbf{I}(i,j)))
\end{equation}
It can be seen that more confident in the prediction of a method is associated with lower entropy. For example, suppose a method has a prediction $P_1=[0.9, 0.05, 0.05]$, then the entropy would be $E_1=0.39$. Another method with prediction $P_2=[0.35, 0.35, 0.3]$, which is less confident than the previous method i.e. the decision is difficult to get from the prediction of the second method, would have entropy $E_2=1.09$. Based on this observation, we define $\theta_k$ as the entropy threshold for $\mathcal{K}_k$. Only the predictions having entropy lower than the corresponding threshold are added into the ensemble. In this way, our approach takes into consideration the confidence of each segmentation algorithm on each pixel:
\begin{equation}
    \label{eq:selection_criteria}
    \begin{cases}
    \textbf{E}_k(\textbf{I}(i,j)) < \theta_k : \mathcal{C}_k \text{ is selected} \\
    \textbf{E}_k(\textbf{I}(i,j)) \geq \theta_k : \mathcal{C}_k \text{ is not selected}
    \end{cases}
\end{equation}
The chosen segmentation algorithms will have their predictions combined via summation:
\begin{equation}
    \label{eq:combine_using_entropy}
    P^{*}_{m}(\textbf{I}(i,j))=\frac{\sum_{k=1}^{K}\mathbb{I}[\textbf{E}_k(\textbf{I}(i,j)) < \theta_k]P_{k,m}(\textbf{I}(i,j))}{\sum_{k=1}^{K}\mathbb{I}[\textbf{E}_k(\textbf{I}(i,j)) < \theta_k])}
\end{equation}
where $P^{*}_{m}(\textbf{I}(i,j))$ is the combined prediction probability for class $y_m$ and $\mathbb{I}[.]$ denotes the indicator function, which is equal to 1 if the condition inside the bracket is true, otherwise it is equal to 0. The class label associated with the maximum value among the combined probabilities is assigned to the pixel $\textbf{I}(i,j)$: 
\begin{equation}
    \label{eq:assign_class}
    \textbf{I}(i,j) \in y_s  \text{ if } s=argmax_{m=1, ... ,M} P_m^{*}(\textbf{I}(i,j))
\end{equation}

We formulate an optimisation problem to find the optimal thresholds $\{\theta_k\}_{k=1}^{K}$ by exploring the ground-truth information of given training data. In this study, we apply the Stacking algorithm to generate the predictions of pixels in training images \cite{nguyen_variational_2016}. The training set $\textbf{D}$ is divided into $T$ disjoint parts $\{\textbf{D}_1,...,\textbf{D}_T\}$, where $\textbf{D}=\textbf{D}_1 \cup ... \cup \textbf{D}_T, \textbf{D}_{t_1} \cap \textbf{D}_{t_2}= \emptyset  (t_1 \neq t_2), |\textbf{D}_1 | \approx ... \approx |\textbf{D}_T |$, and their corresponding remainder $\{\tilde{\textbf{D}}_1,...,\tilde{\textbf{D}}_T\}$ in which ${\tilde{\textbf{D}}}_t=\textbf{D}-\textbf{D}_t$. Each segmentation algorithm $\mathcal{K}_k$ trains on $\tilde{\textbf{D}}_t$ to obtain a model $\textbf{C}_k^{~t}$. Afterwards, $\textbf{C}_k^{~t}$ will segment each image in $\textbf{D}_t$. For a pixel at $(i,j)$ of image  $\textbf{I}$ in the training set  $\textbf{D}$, these models will output a probability vector  $P_{k,m}(\textbf{I}(i,j))$. The predictions for an image $\textbf{I}$ is an $(W \times H) \times (M \times K)$ matrix $\textbf{P}(\textbf{I})$:
\begin{equation}
    \label{eq:pred_each_image}
    \resizebox{0.9\columnwidth}{!}{%
    $\textbf{P}(\textbf{I}) = \begin{bmatrix}
        P_{1,1}(\textbf{I}(1,1)) & \cdots & P_{1,M}(\textbf{I}(1,1)) & \cdots & P_{K,1}(\textbf{I}(1,1)) & \cdots & P_{K,M}(\textbf{I}(1,1)) \\
        
        P_{1,1}(\textbf{I}(1,2))  & \cdots & P_{1,M}(\textbf{I}(1,2)) & \cdots & P_{K,1}(\textbf{I}(1,2)) & \cdots & P_{K,M}(\textbf{I}(1,2)) \\
        
        \vdots & \cdots & \vdots & \cdots & \vdots & \cdots & \vdots \\
        
        P_{1,1}(\textbf{I}(W,H))  & \cdots  &  P_{1,M}(\textbf{I}(W,H))  & \cdots  & P_{K,1}(\textbf{I}(W,H))  & \cdots  & P_{K,M}(\textbf{I}(W,H)) \\
        
    \end{bmatrix}$
    }
\end{equation}
The prediction for all images in the training set $\textbf{D}$ is given by a $(N \times W \times H) \times (M \times K)$ matrix:
\begin{equation}
    \label{eq:pred_training_set}
    \resizebox{0.2\columnwidth}{!}{%
    $\mathcal{P} = \begin{bmatrix}
        \textbf{P}(\textbf{I}_1) \\
        \textbf{P}(\textbf{I}_2) \\
        \cdots \\
        \textbf{P}(\textbf{I}_N) \\
    \end{bmatrix}$
    }
\end{equation}
Next we search for the optimal thresholds $\{\theta_k\}_{k=1}^{K}$ by optimising with respect to a fitness measure. In this study, we use Dice coefficient which is a popular measure to evaluate segmentation results \cite{liu_dice_2019}. 
Let $\textbf{pred}$ and $\textbf{ground}$ denote the final predictions and ground truths of all training pixels:
\begin{align}
    \textbf{pred} &= \{pred_1,pred_2 ... pred_M \} \\
    \textbf{ground} &= \{ground_1,ground_2 ... ,ground_M \}
\end{align}
in which $pred_m$ is the vector of size $(N \times W \times H,1)$ with each element having a value of either 0 or 1 denoting whether the corresponding pixel is predicted to belong to class $y_m$. Likewise $ground_m$ is the vector of size $(N \times W \times H,1)$ associated with the class label $y_m$ which is the ground truth of each pixel in the form of crisp label i.e. belonging to $\{0,1\}$. $ground_m$ is obtained from the ground truth $\{\textbf{Y}_n \}$ while $pred_m$ is obtained based on Equation \ref{eq:combine_using_entropy} and \ref{eq:assign_class} for each row of $\mathcal{P}$. The Dice coefficient associated with the class label $y_m$ is given by:
\begin{equation}
    \label{eq:dice_one_class}
    DC_m=\frac{2 \times pred_m^T ground_m}{||pred_m||^2+||ground_m||^2}
\end{equation}
The average Dice coefficient is the average of all Dice coefficients associated with the class labels.
\begin{equation}
    \label{eq:dice_avg}
    DC=\frac{1}{M} \sum_{m=1}^M DC_m 
\end{equation}
We maximize the average Dice coefficient to find the optimal $\{\theta_k\}_{k=1}^{K}$:
\begin{equation}
    \label{eq:optimisation_problem}
    \begin{aligned}
        \max_{\{\theta_k\}_{k=1}^{K}} \quad & DC \\
        \textrm{s.t.} \quad & 0 \leq \theta_{k} \leq \log M (1 \leq k \leq K)\\
    \end{aligned}
\end{equation}
where the inequality conditions come from the definition of entropy.

In this paper, we use a heuristic approach to solve the optimisation problem. Nowadays, there are many variants of Evolutionary Computation (EC)-based methods, which are inspired by natural processes. An important advantage of these methods compared to classical optimisation algorithms is that they can solve problems having non-differentiable, discontinuous, or multi-modal objective functions which appear in many real-life applications \cite{nguyen_confidence_2020}. Among them, Particle Swarm Optimisation (PSO) is one of the most popular methods. However, a shortcoming of PSO is that it can converge prematurely \cite{liang_comprehensive_2006}. The Comprehensive Learning PSO (CLPSO) \cite{liang_comprehensive_2006} was developed to address this shortcoming by having each particle learn from all particles’ local best position. Specifically, the position $\{{\theta}_k\}^K_{k=1}$ of $i^{th}$ particle will also be associated with a $K$-dimension exemplar vector $e_i=(e_i^1,e_i^2, ... ,e_i^K )$ for comprehensive learning. The exemplar vector is introduced for a particle to learn from the local best $(pbest)$ of itself as well as all the other particles. For example, a particle with the position $(0.13, 0.43, 0.22, 0.74, 0,11)$, the velocity $(0.48, 0.25, 0.52, 0.13$, -$0.15)$, and the exemplar 
$(6, 8, 4, 8, 4)$, would learns/updates the 3rd dimension position value based on the 3rd dimension position value of the 4th particle’s $pbest$.

A particle is assigned randomly with an exemplar vector at initialization. The exemplar will be updated after a number of iterations in which a particle’s $pbest$ does not improve. In order to choose which particle to learn from for each dimension, two random particles are selected and the one with higher fitness value will be assigned as the exemplar for the updated particle on the corresponding dimension \cite{liang_comprehensive_2006, tran_high-dimensional_2019}. Therefore, only one acceleration of constant $c$ is needed. The updated equation for the velocity in the CLPSO is given by:
\begin{equation}
    \label{eq:clpso_update_velocity}
    v_i^k \leftarrow a \times v_i^k+c \times r_1 \times (pbest_{e_i^k}^k-{\theta}_i^k )
\end{equation}
in which $a$ is the inertia weight which controls the velocity speeding rate, $c$ is an acceleration constant used to control the learning rate of the exemplars’ local best, $pbest_{e_i^k}^k$ is the $k^{th}$ dimension of particle’s best position referring to the $k^{th}$ dimension  of exemplar $e_i$, and $r_1$ is a random number drawn from a uniform distribution over $[0,1]$. There are many approaches to setting the inertial weight. In this paper we follow the approach of \cite{liang_comprehensive_2006} in which the inertial weight is updated after each iteration:
\begin{equation}
    \label{eq:inertial_weight_update}
    a(iter) = a_0 * \frac{(a_0-a_1)*iter}{maxIter}
\end{equation}
where $iter$ is the iteration number, $maxIter$ is the maximum number of iterations, $a_0=0.9, a_1=0.4$ and $a(iter)$ is the inertia weight at the current iteration. In CLPSO, when a particle moves out of the search bound, its fitness value and $pbest$ are not updated. Because all exemplars are within range, the particle will eventually return to the search bound. The velocity is also bounded via the following equation:
\begin{equation}
    \label{eq:velocity_bound}
    v^{k}_i=min(v^{k}_{max}, max(v^{k}_{min}, v^{k}_i))
\end{equation}
where $v^{k}_{max}$ and $v^{k}_{min}$ are the maximum and minimum velocity values for $v^{k}_i$. Each particle's position is then updated:
\begin{equation}
    \label{eq:pso_update_position}
    {\theta}_i^k \leftarrow {\theta}_i^k+v_i^k
\end{equation}
Considering that CLPSO has demonstrated state-of-the-art global search capabilities in various applications \cite{hu_comprehensive_2014}, such as optimising reactive power dispatch \cite{mahadevan_comprehensive_2010} and optimising network security \cite{ali_comprehensive_2013}, in this paper we use the CLPSO as the optimisation routine for our proposed method. 


The pseudo-code of the training process of the proposed system is present in Algorithm \ref{algo_training_process}. The inputs to the algorithm consist of the training images and ground truth $\textbf{D}$, $K$ segmentation algorithm $\{\mathcal{K}_k\}_{k=1}^K$, and the CLPSO parameters (the population size $popSize$, the number of iterations $maxIter$, and learning rate controller $c$). $K$  segmentation algorithms $\{\mathcal{K}_k \}_{k=1}^K$ are first trained on $\textbf{D}$ to create models $\{\textbf{C}_k\}_{k=1}^{K}$. Afterwards the prediction $\mathcal{P}$ for all pixels of training images are generated by using the Stacking algorithm (Step 2-8). Algorithm \ref{algo_compute_dice_for_each_candidate} is called for each candidate $\{\theta_k\}_{k=1}^{K}$ generated in the CLPSO to calculate its associated Dice coefficient. In Algorithm \ref{algo_compute_dice_for_each_candidate}, for each row of $\mathcal{P}$ i.e. the predictions of $K$ algorithms for a pixel, the combined probabilities associated with the class labels are calculated by applying Equation \ref{eq:combine_using_entropy} and then a class label for this pixel is assigned by using Equation \ref{eq:assign_class}. On the prediction result for all pixels of $\mathcal{P}$, the final predictions $\textbf{pred}$ can be obtained in the form of crisp labels, then the Dice coefficient can be calculated. The CLPSO runs until it reaches the number of iterations. From the last generation, the candidate  $\{\hat{\theta}_k\}_{k=1}^{K}$ which is associated with the best Dice coefficient is selected as the final solution.

The segmentation process for a test image is described in Algorithm \ref{algo_classification_process}. Given an unsegmented image $\textbf{I}$, we first obtain the predictions $\textbf{P}(\textbf{I})$ for all pixels of $\textbf{I}$ by using the $\{\textbf{C}_k \}_{k=1}^K$ (Step 1).  The $M$ combined probabilities of each pixel then are calculated by using the optimal weight $\{\hat{\theta}_k\}_{k=1}^{K}$ and the predictions (Step 3-4). The Equation \ref{eq:assign_class} is applied to these combined probabilities of this pixel to give the final prediction (Step 5). The predictions for all pixels of $\textbf{I}$ constitute its segmentation result.
\begin{figure}
    \begin{center}
        \includegraphics[width=0.5\textwidth]{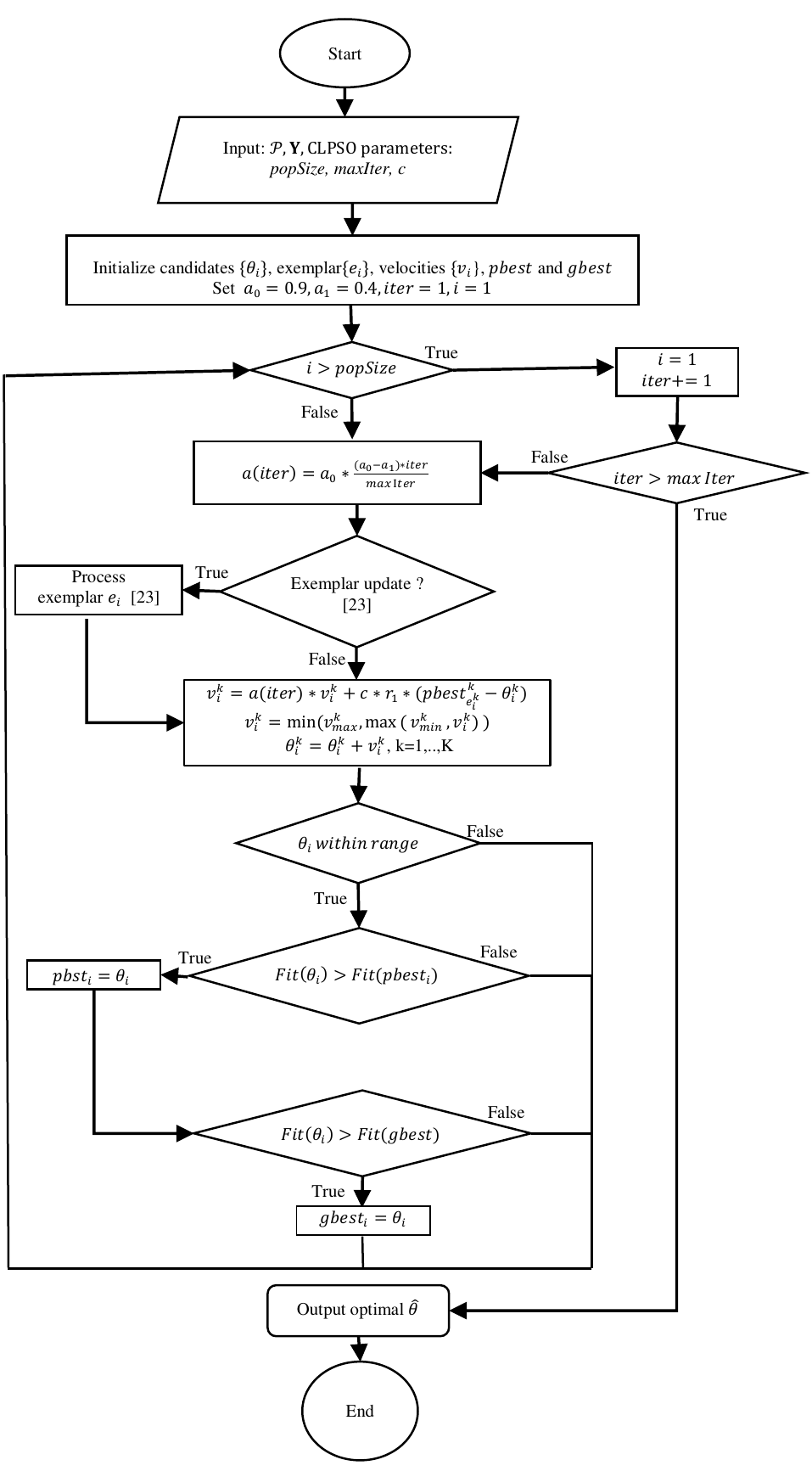}
    \end{center}
    \caption{The flowchart of the CLPSO \cite{liang_comprehensive_2006}}
    \label{fig:diagram_clpso}
\end{figure}
\begin{center}
\label{algo:training_process}
    \begin{algorithm}[t]
    \small
    \caption{\textbf{Training process}}
    \begin{algorithmic}[1]\label{algo_training_process}
        \renewcommand{\algorithmicrequire}{\textbf{Input: }}
        \renewcommand{\algorithmicensure}{\textbf{Output: }}

        \REQUIRE Training images $\textbf{D}$, $K$ segmentation algorithms $\{\mathcal{K}_k \}_{k=1}^K$, parameters for the CLPSO: maximum number of iteration $maxIter$, number of candidates $nPop$, $c$, $a$.\\
        \algrule
        \ENSURE  The optimal weights $\{\hat{\theta}_k\}_{k=1}^{K}$ and segmentation models $\{\textbf{C}_k\}_{k=1}^{K}$ \\
        \algrule

        \STATE Train $K$ models $\{\textbf{C}_k\}_{k=1}^{K}$ on $\textbf{D}$ using $\{\mathcal{K}_k\}_{k=1}^K$
        \STATE $\mathcal{P}= \emptyset$
        \STATE $\textbf{D}=\textbf{D}_1 \cup  ...  \cup \textbf{D}_T, \textbf{D}_{t_1} \cap \textbf{D}_{t_2}= \emptyset (t_1 \neq t_2)$
        \FOR{each $\textbf{D}_t$}
            \STATE $\tilde{\textbf{D}}_t=\textbf{D}-\textbf{D}_t$
            \STATE Train ensemble of segmentation models on $\tilde{\textbf{D}}_t$ using $\{\mathcal{K}_k\}_{k=1}^K$
            \STATE Segment images in $\textbf{D}_t$ by these models
            \STATE Add outputs on samples in $\textbf{D}_t$ to $\mathcal{P}$ (\ref{eq:pred_training_set})
        \ENDFOR
        \STATE Use the CLPSO method \cite{liang_comprehensive_2006}: for each candidate $\{\theta_k\}_{k=1}^{K}$, compute the associated Dice coefficient using Algorithm \ref{algo_compute_dice_for_each_candidate} (see detail in Figure \ref{fig:diagram_clpso})
        \STATE Select the optimal $\{\hat{\theta}_k\}_{k=1}^{K}$ with the best Dice coefficient
        
        \RETURN  $\{\hat{\theta}_k\}_{k=1}^{K}$ and $\{\textbf{C}_k \}_{k=1}^K$

\end{algorithmic}
\end{algorithm}
\end{center}

\begin{center}
\label{algo:compute_dice_for_each_candidate}
    \begin{algorithm}[t]
    \small
    \caption{\textbf{Compute the Dice coefficient for each candidate generated in the CLPSO}}
    \begin{algorithmic}[1]\label{algo_compute_dice_for_each_candidate}
        \renewcommand{\algorithmicrequire}{\textbf{Input: }}
        \renewcommand{\algorithmicensure}{\textbf{Output: }}

        \REQUIRE Candidate $\{\theta_k\}_{k=1}^{K}$, predictions $\mathcal{P}$, ground truth $\{\textbf{Y}_n\}_{n=1}^{N}$\\
        \algrule
        \ENSURE  The Dice coefficient associated with $\{\theta_k\}_{k=1}^{K}$\\
        \algrule

        \FOR{each row $\textbf{I}_n(i,j)$ of $\mathcal{P}$}
            \FOR{$m \gets 1$ to $M$}
                \STATE Compute $P^{*}_{m}(\textbf{I}(i,j))$ by using Equation  \ref{eq:combine_using_entropy} 
            \ENDFOR
            \STATE Assign class label to $\textbf{I}_n(i,j)$ by using Equation  \ref{eq:assign_class}
        \ENDFOR
        \STATE Generate $\textbf{pred}$
        \STATE Generate $\textbf{ground}$
        \STATE Compute $DC$ by Equation  \ref{eq:dice_avg}
        
        \RETURN  $DC$

\end{algorithmic}
\end{algorithm}
\end{center}

\begin{center}
\label{algo:classification_process}
    \begin{algorithm}[t]
    \small
    \caption{\textbf{Segmentation process}}
    \begin{algorithmic}[1]\label{algo_classification_process}
        \renewcommand{\algorithmicrequire}{\textbf{Input: }}
        \renewcommand{\algorithmicensure}{\textbf{Output: }}

        \REQUIRE Unsegmented image $\textbf{I}$, the optimal weights $\{\hat{\theta}_k\}_{k=1}^{K}$ and $\{\textbf{C}_k \}_{k=1}^K$\\
        \algrule
        \ENSURE  Segmented result for $\textbf{I}$\\
        \algrule

        \STATE Obtain the prediction $\textbf{P}(\textbf{I})$ by using $\{\textbf{C}_k \}_{k=1}^K$
        
        \FOR{each pixel of $\textbf{I}$}
            \FOR{$m \gets 1$ to $M$}
                \STATE Compute $P^{*}_{m}(\textbf{I}(i,j))$ by using Equation \ref{eq:combine_using_entropy} with $\{\hat{\theta}_k\}_{k=1}^{K}$
                \ENDFOR
            \STATE Assign label to $\textbf{I}(i,j)$ by using Equation  \ref{eq:assign_class}
        \ENDFOR

        \RETURN Segmented result for $\textbf{I}$

\end{algorithmic}
\end{algorithm}
\end{center}
\sloppy

\section{Experimental Studies} \label{sec:experimental_studies}
In this experiment, we used three popular deep learning-based segmentation methods UNet \cite{unet_olaf_2015}, LinkNet \cite{linknet_paper_2017} and Feature Pyramid Network (FPN) \cite{fpn_paper_2017} with three backbone  VGG16 \cite{vgg16_paper_2015},  ResNet34 and ResNet101 \cite{resnet_paper_2016} to create an ensemble of $K=9$ segmentation algorithms. Thus the number of search dimensions is $K=9$. These backbones were pretrained on the ImageNet dataset \cite{imagenet_cvpr_2009}. All segmentation algorithms were run for 300 epochs. The 5-fold cross-validation was used in the experiments and was run using GPU. The performance of the proposed ensemble was compared to those of these 9 segmentation algorithms and those of datasets' authors. The CLPSO algorithm was run for 500 iterations on the Core i5 CPU. The number of candidates $nPop$ used for the CLPSO search was set to 10. Dice coefficients of all algorithms were report with the note that a high Dice coefficient is an indication of good segmentation result.
\subsection{MICCAI 2015 dataset}

The first dataset in our experiment is from the MICCAI 2015 Endoscopic Vision Challenge \cite{miccai_2015_paper}, which is a challenge for colorectal polyp detection and localisation. The dataset contains 612 training images and 196 test images. Each image contains at least one polyp and have been selected in order to have shots in which polyp appearance can be mistaken with other elements of the scene. There are two classes: polyp and background. Creating the prediction matrix $\mathcal{P}$ in Equation \ref{eq:pred_training_set} for the optimisation routine took 2 days while the CLPSO was run for 500 iterations, taking 1.5 days. Table \ref{tab:result_miccai2015} shows the results from the winning solutions reported in the challenge: CUMED \cite{miccai_2015_paper}, CVC-CLINIC \cite{bernal_wm-dova_2015}, ETIS-LARIB \cite{silva_toward_2014}, OUS , PLS, SNU and UNS-UCLAN (the results of these methods were not published in a paper), the 9 segmentation benchmarks mentioned in Section  \ref{sec:experimental_studies} and the proposed ensemble. It can be seen that the proposed ensemble achieves the best Dice coefficient for all three classes compared to all other benchmarks. Among the authors' mentioned methods, CUMED had the best Dice coefficient at 0.707, followed by OUS at 0.661. Other methods achieved much lower scores from 0.099 (SNU) to 0.404 (UNS-UCLAN). The proposed ensemble scored 0.724 with respect to the Dice coefficient, which is 1.7\% higher than CUMED. This is followed by FPN-ResNet101, which was at 0.682. The networks using ResNet34 backbone scored around 0.53 to 0.58, while UNet-ResNet101 and LinKNet-ResNet101 achieved only 0.487 and 0.46 respectively. The VGG16-based methods achieved very low Dice coefficients, less than 0.1. 

\begin{figure}
    \begin{center}
        \includegraphics[width=0.45\textwidth]{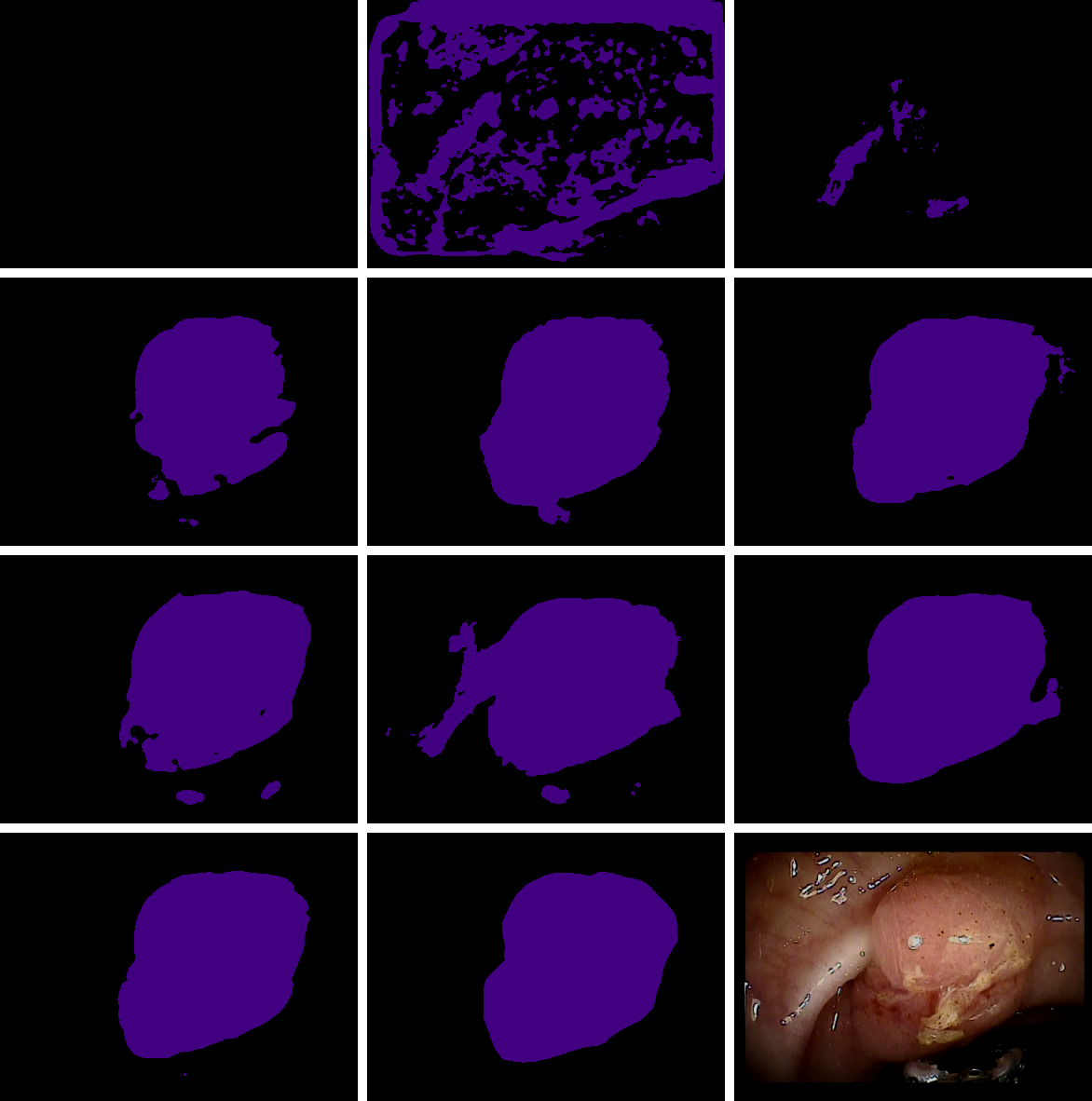}
    \end{center}
    \caption{Example result from MICCAI2015. From left to right, top to bottom: UNet-VGG16, LinkNet-VGG16, FPN-VGG16, UNet-ResNet34, LinkNet-ResNet34, FPN-ResNet34, UNet-ResNet101, LinkNet-ResNet101, FPN-ResNet101, proposed ensemble, ground truth, test image}
    \label{fig:miccai2015_example_result}
\end{figure}

Figure \ref{fig:miccai2015_example_result} shows an example in which the proposed ensemble provides a better result compared to those of segmentation algorithms. It can be seen that the predictions by the VGG16-based methods (first row) were completely spurious, while the other methods were able to predict the general shape of the polyp but had a number of defects, which the proposed ensemble had corrected. Both UNet-ResNet34 and UNet-ResNet101 (first column, second and third row respectively) did not correctly segment the bottom left of the polyp, and their predictions also contained a number of internal areas that were not segmented. The other four segmentation algorithms predicted the general shape of the polyp, but wrongly segmented several adjacent areas as polyp. Specifically, the prediction by LinkNet-ResNet34 (second row, second column) had a redundant part in the bottom. This can also be observed with the case of LinkNet-ResNet101 (third row, second column) although unlike with LinkNet-ResNet34 this part is now separated from the main polyp. There is also a large redundant area on the left and a small area on the right. The predictions of FPN-ResNet34 (second row, third column) and FPN-ResNet101 (third row, third column) had a redundant area in the top right and bottom right part respectively. In contrast, the segmentation by the proposed ensemble (fourth row, first column) is very similar to the ground truth (fourth row, second column), although it can be seen that the left curve of the polyp is slightly less inward in the proposed ensemble's prediction as compared to the ground truth.

\begin{table}[t]
\centering
\caption{Result for the MICCAI2015 datasets. CUMED, CVC-CLINIC, ETIS-LARIB, OUS, PLS, SNU and UNS-UCLAN were the winning solutions reported in the challenge \cite{miccai_2015_paper}, while UNet-VGG16, LinkNet-VGG16, FPN-VGG16, UNet-ResNet34, LinkNet-ResNet34, FPN-ResNet34, UNet-ResNet101, LinkNet-ResNet101 and FPN-ResNet101 were the base segmentation methods used in the proposed ensemble.}
\label{tab:result_miccai2015}
\resizebox{0.45\columnwidth}{!}{
\begin{tabular}{@{}lllllll@{}}
\toprule
Method                                & Dice \\  \midrule
CUMED \cite{miccai_2015_paper}       & 0.707           \\
CVC-CLINIC \cite{bernal_wm-dova_2015}      & 0.165                  \\
ETIS-LARIB  \cite{silva_toward_2014}     & 0.122                      \\
OUS \cite{miccai_2015_paper}      & 0.661                     \\
PLS \cite{miccai_2015_paper}      & 0.249                   \\
SNU \cite{miccai_2015_paper}      & 0.099                     \\
UNS-UCLAN \cite{miccai_2015_paper}      & 0.404                       \\
UNet-VGG16 & 0.006   \\ 
LinkNet-VGG16 & 0.071   \\ 
FPN-VGG16 & 0.028  \\ 
UNet-ResNet34 & 0.538  \\ 
LinkNet-ResNet34 & 0.581   \\ 
FPN-ResNet34 & 0.561   \\ 
UNet-ResNet101 & 0.487   \\ 
LinkNet-ResNet101 & 0.46   \\ 
FPN-ResNet101 & 0.682  \\ 
Proposed ensemble & \textbf{0.724}  \\
\bottomrule
\end{tabular}}
\end{table}
\subsection{Kvasir-SEG dataset}
\begin{figure}
    \begin{center}
        \includegraphics[width=0.45\textwidth]{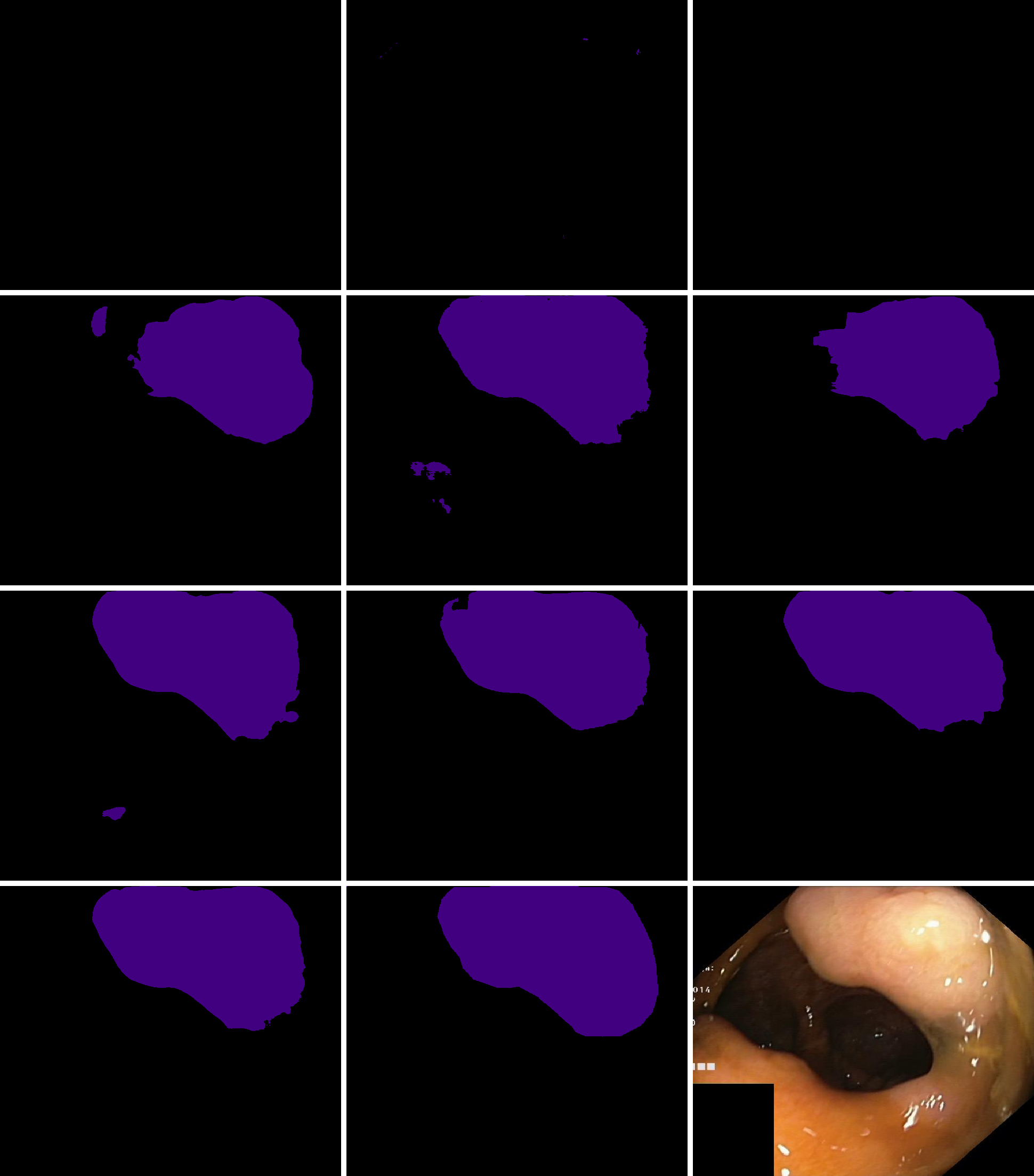}
    \end{center}
    \caption{Example result from Kvasir-SEG.  From left to right, top to bottom: UNet-VGG16, LinkNet-VGG16, FPN-VGG16, UNet-ResNet34, LinkNet-ResNet34, FPN-ResNet34, UNet-ResNet101, LinkNet-ResNet101, FPN-ResNet101,  proposed ensemble, ground truth, test image}
    \label{fig:kvasir_seg_example_result}
\end{figure}

The second dataset used in this paper is Kvasir-SEG \cite{jha_kvasir-seg_2020}, an open-access dataset of gastrointestinal polyp
images and corresponding ground truths, annotated by a medical doctor and verified by an experienced gastroenterologist. This dataset was based on the Kvasir dataset, which is the multi-class dataset for gastrointestinal (GI) tract disease detection and classification. 1000 images having polyp were extracted from this dataset to create the Kvasir-SEG dataset, 200 of which were used for testing. It took 4 days to run the $5$-fold cross-validation to create the prediction matrix $\mathcal{P}$ in \ref{eq:pred_training_set}, and running 500 iterations of CLPSO took 2 days. Table \ref{tab:result_kvasir_seg} summarizes the results by the authors, the segmentation benchmarks and the proposed ensemble. From the results, it can be seen that the proposed ensemble achieved the best result, while those with the VGG16 backbone obtained very low scores. The proposed ensemble scored the highest in the Dice coefficient at 0.894, followed by ResNet101-based methods which scored from 0.886 (LinkNet-ResNet101) to 0.89 (UNet-ResNet101). The author's result was only 0.787763. The methods using the VGG16 backbone had very low scores at just around 0.001. Figure \ref{fig:kvasir_seg_example_result} shows an example of predictions by the segmentation algorithms, the proposed ensemble compared with the ground truth. The segmentation algorithms with VGG16 backbone (first row) could not segment the polyp, while UNet-ResNet34 (second row, first column) could not segment the left area of the polyp except for a small part. Similarly, in the case of FPN-ResNet34 (second row, third column), both the left part and the bottom right part were not segmented. Both LinkNet-ResNet34 (second row, second column) and UNet-ResNet101 (third row, first column) wrongly considered a small isolated area in the bottom left as part of the polyp. There is also a small spike in the bottom right of the polyp prediction by UNet-ResNet101. LinkNet-ResNet101 and FPN-ResNet101 have better predictions compared to the previously mentioned segmentation algorithms, however they did not segment a small region in the top left and bottom right respectively. It can be seen that the proposed ensemble has provided an improved prediction compared to the benchmarks, even though there is a small unsegmented region in the bottom right which was also not segmented by all the benchmarks. Another important point is that this additional gain in both datasets was achieved at a small computational cost.


\begin{table}[t]
\centering
\caption{Result for the Kvasir-SEG datasets}
\label{tab:result_kvasir_seg}
\resizebox{0.5\columnwidth}{!}{
\begin{tabular}{@{}lllllll@{}}
\toprule
Method                                    & Dice \\  \midrule
Author's result       & 0.787763           \\
UNet-VGG16 & 0   \\ 
LinkNet-VGG16 & 0.001   \\ 
FPN-VGG16 & 0  \\ 
UNet-ResNet34 & 0.878 \\ 
LinkNet-ResNet34 & 0.879 \\ 
FPN-ResNet34 & 0.887  \\ 
UNet-ResNet101 & 0.89 \\ 
LinkNet-ResNet101 & 0.886  \\ 
FPN-ResNet101 & 0.887   \\ 
Proposed ensemble & \textbf{0.894} \\ 
\bottomrule
\end{tabular}}
\end{table}

\subsection{Entropy threshold}
\begin{figure}
    \begin{center}
        \includegraphics[width=0.46\textwidth]{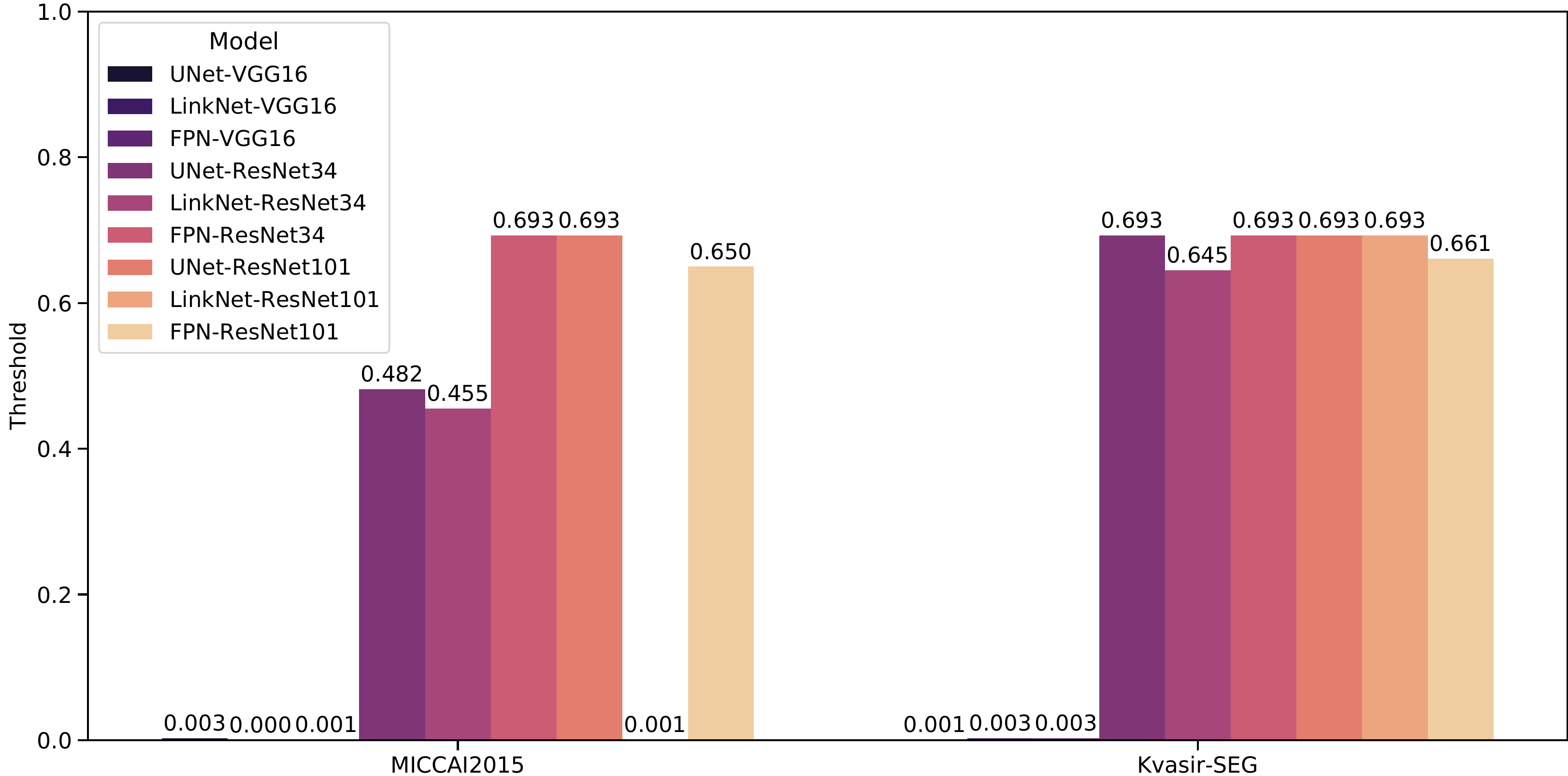}
    \end{center}
    \caption{Shannon entropy thresholds found by using CLPSO for two datasets MICCAI2015 and Kvasir-SEG}
    \label{fig:entropy_threshold_graph}
\end{figure}
Figure \ref{fig:entropy_threshold_graph} shows the entropy thresholds of all methods found by using CLPSO on two datasets MICCAI2015 and Kvasir-SEG. It can be seen that the methods having the VGG16 backbone were assigned very low thresholds (around 0.001) compared to other methods on both datasets. This is because the VGG16-based methods performed poorly, therefore they were only selected if very confident about their prediction. The small thresholds show that these methods were rarely selected to the ensemble. In contrast, other high-performing methods were likely to be selected even when they are mildly confident about their predictions. For the MICCAI2015 dataset, FPN-ResNet34 and UNet-ResNet101 had the maximum threshold at $log M=0.693 (M=2)$. These methods thus were always selected to the ensemble (see Equation \ref{eq:selection_criteria}). The threshold of FPN-ResNet101 was 0.650 while UNet-ResNet34 and LinkNet-ResNet34 only had the thresholds of 0.482 and 0.455 respectively. LinkNet-ResNet101 had very low entropy threshold (0.001) because it achieved a Dice coefficient of only 0.46 while other ResNet-based methods achieve from around 0.48 to around 0.68 (Table \ref{tab:result_miccai2015}). With respect to the Kvasir-SEG dataset, most ResNet34-based methods had the highest entropy threshold at 0.693 with the exception of LinkNet-ResNet34 (0.645) and FPN-ResNet101 (0.661), which can be explained by the fact that most segmentation algorithms achieved similar Dice coefficient from 0.878 to 0.887 (Table \ref{tab:result_kvasir_seg}). It can also be seen that the optimal thresholds depend on the dataset as well as the performance of each method.

\section{Conclusion}
In this paper, we presented an ensemble of medical polyp segmentation algorithms. Our approach takes into consideration the fact that the presence of some segmentation algorithms might degrade ensemble performance, thus needing to remove from the ensemble. We introduced a novel ensemble selection method. The key idea is to measure uncertainty in the prediction of each model. If the uncertainty is below its associate threshold, the prediction is confident and it is selected to calculate the combined prediction. Shannon entropy is used as the uncertainty measure. The optimal entropy threshold for each segmentation algorithm is found by using Comprehensive Learning Particle Swarm Optimisation (CLPSO), a swarm intelligence algorithm. Dice coefficient, which is a popular performance metric for image segmentation, is used as the fitness criteria. Our experiments on two polyp segmentation datasets, MICCAI2015 and Kvasir-SEG, show that the proposed ensemble provides better results compared with some well-known segmentation algorithms. The use of CLPSO can obtain different thresholds for constituent algorithms of the ensemble.

\bibliographystyle{ACM-Reference-Format}
\bibliography{bibfile.bib}

\end{document}